\documentclass[10pt,twocolumn,letterpaper]{article}

\usepackage{cvpr}
\usepackage{times}
\usepackage{epsfig}
\usepackage{graphicx}
\usepackage{amsmath}
\usepackage{amssymb}
\usepackage{array}
\usepackage{subfigure}
\usepackage{flushend}
\usepackage{authblk}

\usepackage[breaklinks=true,bookmarks=false]{hyperref}

\cvprfinalcopy 

\def\cvprPaperID{****} 

\begin{document}

\title{DF-SLAM: A Deep-Learning Enhanced Visual SLAM System based on Deep Local Features}
\author[1]{Rong Kang$^*$}
\author[2]{Jieqi Shi\thanks{The first two authors contribute equally to this paper.}}
\author[1]{Xueming Li}
\author[1]{Yang Liu}
\author[2]{Xiao Liu}
\affil[1]{School of Digital Multimedia and Design Arts, Beijing University of Posts and Telecommunications, China}
\affil[2]{Megvii Co. \authorcr \{kangr, lixm, yang.liu\}@bupt.edu.cn, \{shijieqi, liuxiao\}@megvii.com}
\maketitle

\maketitle

\begin{abstract}
	As the foundation of driverless vehicle and intelligent robots, Simultaneous Localization and Mapping(SLAM) has attracted much attention these days. However, non-geometric modules of traditional SLAM algorithms are limited by data association tasks and have become a bottleneck preventing the development of SLAM. To deal with such problems, many researchers seek to Deep Learning for help. But most of these studies are limited to virtual datasets or specific environments, and even sacrifice efficiency for accuracy. Thus, they are not practical enough.
	
	We propose DF-SLAM system that uses deep local feature descriptors obtained by the neural network as a substitute for traditional hand-made features. Experimental results demonstrate its improvements in efficiency and stability. DF-SLAM outperforms popular traditional SLAM systems in various scenes, including challenging scenes with intense illumination changes. Its versatility and mobility fit well into the need for exploring new environments. Since we adopt a shallow network to extract local descriptors and remain others the same as original SLAM systems, our DF-SLAM can still run in real-time on GPU.
\end{abstract}

\section{Introduction}

Traditional SLAM(Simultaneous Localization and Mapping) systems paid great attention to geometric information. Based on the solid foundation of Multi-view Geometry, a lot of excellent studies have been carried out. However, problems arise from none-geometric modules in SLAM systems. To track the location of cameras, researchers usually perform pixel-level matching operations in tracking threads and optimize poses of a small number of frames as local mapping. No doubt that errors resulted by drift in pose estimation and map evaluation keep accumulating.

In the meanwhile, Deep Learning, a data-driven technique, has brought out rapid development in numerous computer vision tasks such as classification and matching. Such achievements reflect that deep learning may be one of the best choices to solve problems related to data association. Therefore, more and more researchers believe that pixel-level or higher level associations between images, the bottleneck of SLAM systems we mentioned above, can also be handled with the help of neural networks.

Deep learning has proved its superiority in SLAM systems. Many outstanding studies have employed it to replace some non-geometric modules in traditional SLAM systems \cite{kottas2013efficient,kong2015tightly,zbontar2016stereo,luo2016efficient,feng2017efficient}. These approaches enhance the overall SLAM system by improving only part of a typical pipeline, such as stereo matching, relocalization and so on. Some researchers also attempt to use higher-level features obtained through deep learning models as a supplement to SLAM \cite{salas2013slam++,reid2014towards,atanasov2014semantic,bowman2017probabilistic,gay2017probabilistic}.These higher-level features are more likely to infer the semantic content-object feature and improve the capability of visual scene understanding. Moreover, end-to-end learning models have also been proposed\cite{zhu2017target,gupta2017cognitive}. These methods outperform traditional SLAM algorithms under specific circumstances and demonstrate the potential of deep learning in SLAM.

However, such combination of Deep learning and SLAM have significant shortcomings. Most of  Deep Learning methods rely heavily on data used for training, which means that they can not fit well into unknown environments. For example, we can not ensure whether the room we want to explore is equipped with chairs and desks and cannot guarantee semantic priority of desks will help in this occasion. What's more, most Deep-Learning enhanced SLAM systems are designed to reflect advantage of Deep Learning techniques and abandon the strong points of SLAM. As a result, they may sacrifice efficiency, an essential part of SLAM algorithms, for accuracy.

Last but not least, some DL-based SLAM techniques take traditional SLAM systems as their underlying framework\cite{zbontar2016stereo,luo2016efficient,feng2017efficient,detone2017superpoint} and make a great many changes to support Deep Learning strategies. Too many replacements may lead to loss of some useful features of the SLAM pipeline and also make it hard for researchers to perform further comparisons with existing studies, let alone migrate these techniques to other SLAM systems. As a result, DL-based SLAM is not mature enough to outperform traditional SLAM systems.

Therefore, we make our efforts to put forward a simple, portable and efficient SLAM system. Our basic idea is to improve the robustness of local feature descriptor through deep learning to ensure the accuracy of data association between frames.

In this paper, we propose a novel approach to use the learned local feature descriptors as a substitute for the traditional hand-craft descriptors. Our method has advantages in portability and convenience as deep feature descriptors can directly replace traditional ones. The replacement is highly operable for all SLAM systems and even other geometric computer vision tasks such as Structure-from-Motion, camera calibration and so on. The learned local feature descriptors guarantee better performance than hand-craft ones in actual SLAM systems \cite{mur2017orb} and achieve amazing improvement in accuracy. Since we adopt a shallow neural network to obtain local feature descriptor, the feature extraction module does not consume much time on GPU, and the system can operate in almost real-time.

\section{Related Work}

\textbf{Deep Learning enhanced SLAM.} Deep learning is considered an excellent solution to SLAM problems due to its superb performance in data association tasks. Part of recent studies makes a straight substitution of an end-to-end network for the traditional SLAM system, estimating ego-motion from monocular video\cite{zhou2017unsupervised,mahjourian2018unsupervised,li2018undeepvo} or completing visual navigation for robots entirely through neural networks\cite{zhu2017target,gupta2017cognitive}. Such works can hardly catch up with traditional methods in accuracy under test datasets. Nevertheless, since deep learning systems rely too much on training data, the end-to-end system fails from time to time at the face of new environments and situations. That's to say the model may hardly predict correct results when there exists a big difference between training scenes and actual scenes.

To tackle such problems, some researchers focus on the replacement of only parts of traditional SLAM systems while keeping traditional pipelines unchanged\cite{garg2016unsupervised,yang2018deep}\cite{kendall2015posenet,wu2017delving,vo2017revisiting}. Such attempts are still in an embryonic stage and do not achieve better results than traditional ones. One of the possible explanation for their limited improvement is that they also rely too much on the priority learned from training data, especially when it comes to predicting depth from monocular images. Thus, they are still subject to the same limitation of end-to-end methods. We believe that the experience-based system is not the best choice for geometric problems.

Other efforts are made to add auxiliary modules rather than replace existing geometric modules.  Semantic mapping and fusion\cite{reid2014towards,semanticfusion} make use of semantic segmentation. They always take in poses provided by underlying SLAM systems and output optimized 3D models. Such changes are not involved in the optimization of original SLAM systems and cannot directly improve pose estimation modules. Some other researchers separate key points belonging to different items and process them differently \cite{engel2018direct}. These constraints have outstanding performance especially when the environment is dynamic. But they still avoid making changes to the basic system. To combine higher-level information tighter with SLAM pipelines, Detection SLAM and Semantic SLAM\cite{salas2013slam++} jointly optimize semantic information and geometric constraints. Early studies operate semantic and geometric modules separately and merge the results afterward\cite{civera2011towards,pillai2015monocular}. \cite{atanasov2014semantic} incorporate semantic observations in the geometric optimization via Bayes filter. Focusing on the overall SLAM pipeline, \cite{bowman2017probabilistic,gay2017probabilistic} formulate semantic SLAM as a probabilistic model. These approaches extract object-level information and add the semantic feature to the constraints of Bundle Adjustment. However, up to now, there are still no convincing loss functions for semantic modules, and there are also no outbreaking improvements. What's worse, since semantic SLAM add too much extra supervision to the traditional SLAM systems, the number of variables to be optimized inevitably increased, which is a great challenge for the computation ability and the speed.

A simple but effective method is to directly improve the module that limits the performance of traditional SLAM, i.e., stereo matching between frames. Some of them calculate similarity confidence of local features\cite{zbontar2016stereo,luo2016efficient,feng2017efficient}, resulting in the inability to use traditional matching strategy, such as Euclidean distance, cosine distance and so on. SuperPoint\cite{detone2017superpoint} trains an end-to-end network to extract both local feature detectors and descriptors from raw images through one forward calculation. However, the efficiency of SuperPoint remains not verified as it only gives out the result on synthetic and virtual datasets and has not been integrated into a real SLAM system for evaluation.

\textbf{Local feature descriptor.} Parallel with the long history of SLAM, considerable attempts have been made on local features. Based on classical hand-craft local features like SIFT \cite{ng2003sift}, SURF \cite{bay2006surf}, ORB \cite{rublee2011orb}, early combination of low-level machine learning and local feature descriptors produce PCA-SIFT \cite{ke2004pca}, ASD \cite{wang2014affine}, BOLD \cite{balntas2015bold}, Binboost \cite{trzcinski2013boosting}, RFD \cite{fan2014receptive}, RMGD \cite{gao2015local}, GRIEF\cite{krajnik2016griefras} etc. Some of these attempts dedicate on dimensionality reduction and utilize various methods to map high-dimensional descriptors to low-dimensional space. Thus they lose a great amount of information on the raw image. Others make use of binary features. Part of them enhance a traditional feature on specific environments to fit special requirements\cite{krajnik2016griefras} and is lack of mobility. Most of these studies put forward a new kind of feature without further tests or applications.

Thanks to the booming of Deep Learning, researchers have gone further. End-to-end networks consisting of multiple independent components\cite{yi2016lift,detone2017superpoint,ono2018lf,noh2017largescale} can not only give out local feature descriptors through one forward computation but also extract local feature detectors.

Focusing only on descriptors, most researchers adopt multi-branch CNN-based architectures like Siamese and triplet networks. Multi-branch networks were first proposed to verify whether the handwritten signatures were consistent in 1994 \cite{bromley1994signature}. Experiments related to similarity measurements further confirm the superiority of this multi-branch structure. As a result, Siamese and triplet networks turn out to be the main architectures employed in local feature descriptor tasks. MatchNet\cite{han2015matchnet} and DeepCompare\cite{zagoruyko2015learning} are typical Siamese networks. Each branch consists of a feature network and a metric network which determines the similarity between two descriptors. Thus the final output is similarity confidence. Together with the metric learning layer, \cite{kumar2016learning} uses triplet structure and achieves better performance. These achievements reveal the potential of triplet neural network. However, these models prove to be not suitable for traditional nearest neighbor search. Therefore, studies that directly output local feature descriptors are derived.

Early research\cite{simo2015discriminative} only uses Siamese network and designs a novel sampling strategy. L2Net \cite{tian2017l2} creatively utilizes a central-surround structure and a progressive sampling strategy to improve performance. These unique structures and training strategies can also extend to triplet. \cite{mishchuk2017working} adopts the structure presented by L2Net and enhances the strict hardest negative mining strategy to select closest negative example in the batch. \cite{he2018local} also uses the same structure but formulates feature matching as nearest neighbor retrieval. Thus, it directly optimizes a ranking-based retrieval performance metric to obtain the model. It is worth to be mentioned that \cite{balntas2016learning} trains a shallow triplet network based on random sampling strategy but performs better than some deep structures like DeepDesc and DeepCompare, which is an essential reference for our work.  Similar to TFeat, some researchers focus on the formation of a single branch. DeepCD \cite{yang2017deepcd} proposes a new network layer, termed the data-dependent modulation layer, to enhance the complementarity of local feature descriptors. Considering that the geometric repeatability is not the only factor that influence learned local features, AffNet \cite{via2017repeatability} raises a novel loss function and training process to estimate the affine shape of patches. It trains local feature descriptor network based on the affine invariance to improve the performance of deep descriptor.
\section{System Overview}

In our DF-SLAM system, learned local feature descriptors are introduced to replace ORB, SIFT and other hand-made features. We adopt the traditional and popular pipeline of SLAM as our foundation and evaluate the efficiency and effectiveness of our improved deep-feature-based SLAM system. The whole system incorporates three threads that run in parallel: tracking, local mapping and loop closing. Local feature descriptors are extracted as long as a new frame is captured and added before the tracking thread.

\subsection{System Framework}
\begin{figure}[h]
	\centering
	\fbox{
		\includegraphics[width=0.95\linewidth]{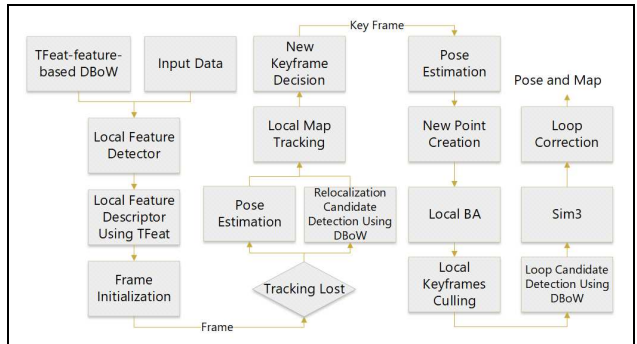}
	}
	\caption{System framework.}
	\label{img}
\end{figure}

The framework of our system is shown in Fig.1. We derive the tracking thread from Visual Odometry algorithms. Tracking takes charge of constructing data associations between adjacent frames using visual feature matching. Afterward, it initializes frames with the help of data associations and estimates the localization of the camera using the polar geometric constraint. It also decides whether new keyframes are needed. If lost, global relocalization is performed based on the same sort of features. Local Mapping will be operated regularly to optimize camera poses and map points. It receives information constructed by the tracking thread and reconstructs a partial 3D map. If loops are detected, the Loop Closure thread will take turns to optimize the whole graph and close the loop. The frame with a high matching score is selected as a candidate loop closing frame, which is used to complete loop closing and global optimization. None of these modules accept raw images as inputs to reduce space consumption. Only sparse visual features and inter-frame associations are recorded to support pose estimation, relocalization, loop detection, pose optimization and so on. Therefore, we believe that the local feature is the cornerstone of our entire system.

As is shown in Fig.2, our first step is to extract our interested points. We utilize TFeat network to describe the region around key points and generate a normalized 128-D float descriptor. Different from hand-made features, we do not need a Gaussian Blur before feature-extraction but take patches of raw images as our input directly. Features extracted are then stored in every frame and passed to tracking, mapping and loop closing threads.

We adopt the method used in ORB-SLAM to perform localization based on DBoW. This method measures the similarity between two frames according to the similarity between their features. As the deep feature descriptor is a float, the Euclidean distance is used to calculate the correspondence. Apparently, the relocalization and loop closing modules rely heavily on the local feature descriptors.

To speed up the system, we also introduce our Visual Vocabulary. Visual Vocabulary is employed in numerous computer vision applications. It extracts a big set of descriptors from training sets offline and creates a vocabulary structured as a tree. Descriptors are divided and integrated according to their characteristics. Thus, during the matching step, a new descriptor could search along the tree for its class much more quickly while ensuring accuracy, which is ideal for practical tasks with real-time requirements. We trained the vocabulary, based on DBoW, using the feature descriptors extracted by our DF methods. Therefore, we could assign a word vector and feature vector for each frame, and calculate their similarity more easily.
\begin{figure}[h]
	
	\centering
	\fbox{
		\includegraphics[width=0.9\linewidth]{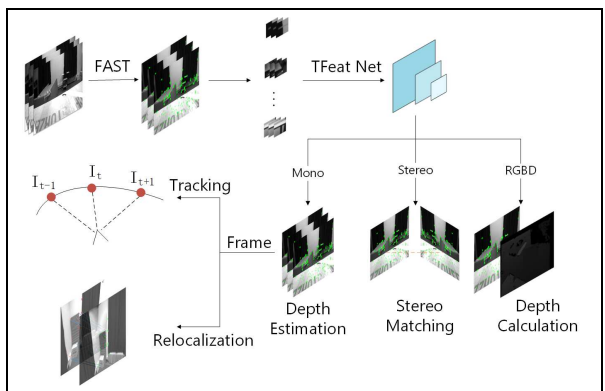}
	}
	\caption{Overview of feature-based modules.}
	\label{img}
\end{figure}

\subsection{Feature Design}

\begin{figure*}[ht]
	\centering
	\fbox{
		\includegraphics[width=0.8\linewidth]{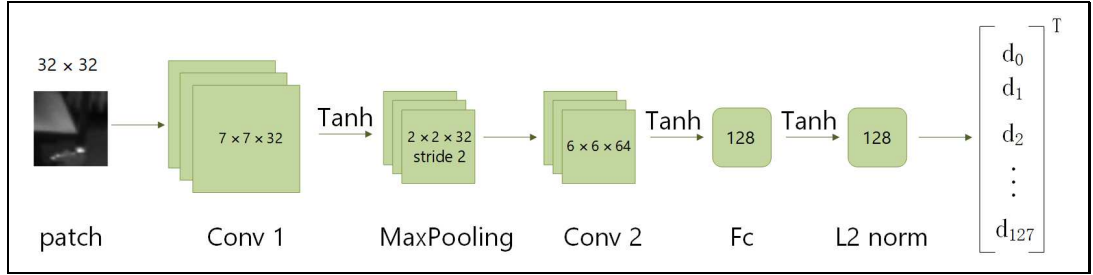}
	}
	\caption{The architecture of TFeat.}
	\label{img}
\end{figure*}

Many excellent studies have indicated the effectiveness of CNN-based neural networks in local feature descriptor designs. However, it's a question of striking the right balance between efficiency and accuracy. Although the performance becomes better and better as the number of convolutional layers increases, time assumption prevents us from adopting a deep and precise network. Instead, we make use of a shallow but efficient network to complete our task.

The architecture adopts a triplet network proposed by TFeat\cite{balntas2016learning}. There are only two convolutional layers followed by Tanh non-linearity in each branch. Max pooling is added after the first convolutional layer to reduce parameters and further speed up the network. A fully connected layer outputs a 128-D descriptor L2 normalized to unit-length as the last layer of the network.

\cite{balntas2016learning} forms triplets for training based on simple methods. It randomly chooses a positive pair of patches that originate from the same label and a sampled patch from another different label. This training strategy is too naive and can hardly improve the performance of the model. Luckily, the hard negative mining strategy proposed in HardNet\cite{mishchuk2017working} is proved to be useful in experiments. We turned to it for help and combined hard negative mining strategy with TFeat architecture to make improvements\footnote{The combination is mentioned in HardNet and AffNet.}.

The sampling strategy selects the closest non-matching patch in a batch by L2 pairwise distance matrix\footnote{The strategy is utilized in HardNet.}. The first step is to generate a batch of matched local patches. Such patches follow the rule that there is only one matching patch for the specific anchor in a batch. To evaluate the similarity of patches, we denote the distance matrix as $D = \{d_{ij}\}$. Each element represents the distance between the $i$th anchor patch descriptor and the $j$th positive patch descriptor.

Next, the hardest negative patch distance can be calculated according to the following rules:
\[
\begin{split}
d_n = min(a_{k_{min}},p_{j_{min}})
\end{split}
\]
where $a_{k_{min}}$ represents the nearest patch to anchor and $p_{j_{min}}$ is the nearest one to positive.

Loss function is formulated as
\[
\begin{split}
Loss = &\frac{1}{N}\sum_{i=0}^{N}max(0,1+d(a_i,p_i)-d_n)
\end{split}
\]

where $a_i$ is anchor descriptor and $p_i$ is positive descriptor.

\section{Experiments}
We perform several experiments to evaluate the efficiency and accuracy of our system and provide some quantitative results. To give out an intuitive comparison, we choose the open-source library of ORB-SLAM as our basis and test on public datasets. All the experiments are performed on a computer with Intel Core i5-4590 CPU 3.30GHz * 4 and GeForce GTX TITAN X/PCIe/SSE2 processor.

\subsection{Preprocess}

Two of the most complicated preparations we made is to create datasets for model training and to construct our visual vocabulary.

Most of the existing patch-based datasets use the DoG detector to extract points of interest. However, the local feature used in most SLAM systems are extracted by a FAST detector and evenly distributed across the image. To fit the requirements of SLAM systems, we need to build patch datasets for training in the same way as ORB-SLAM to ensure the efficiency of the network. We extract our patch from HPatches images containing 116 scenes\cite{balntas2017hpatches}. The patch generation approaches are identical to HPatches except for the way of local feature detection.

After we have successfully received our model, we start another training procedure for visual vocabulary. We train our bag of words on COCO datasets and choose 1e6 as the number of leaves in the vocabulary tree. Since our descriptor is a normalized float vector, the leaf nodes are also normalized.
\subsection{System Performance}

We evaluate the performance of our system in two different datasets to show how well our system can fit into different circumstances. Since we never train our model on these validation sets, the experiments also reveal the modality of our system.

Note that there are many parameters, including knn test ratio in feature matching, number of features, frame rate of camera and others in the original ORB-SLAM2 system. To ensure fairness, we use the same sort of parameters for different sequences and datasets. Such behavior also illustrates how robust and portable our system is.

\subsubsection*{A. EuRoC Dataset}

\begin{table}[h]
	\centering
	\linespread{1}
	\begin{tabular}{cccc}
		\hline
		Dataset & ORB-SLAM2  & DF-SLAM & Improvement \\
		\hline
		MH\_01  & 0.036 & 0.037 & 1.67\%\\
		\hline
		MH\_02  & 0.048 & 0.043 & 10.2\%\\
		\hline
		MH\_03 & 0.044 & 0.046 & -4.9\%\\
		\hline
		MH\_04 & 0.112 & 0.063 & 43.9\%\\
		\hline
		MH\_05 & 0.061 &0.042 & 30.7\%\\
		\hline
		V1\_01 & 0.087 &0.086 & 0.6\%\\
		\hline
		V1\_02 & 0.065 &0.064 & 1.2\%\\
		\hline
		V1\_03 & 0.078 &0.065 & 15.5\%\\
		\hline
		V2\_01 & 0.062 &0.058 & 7.3\%\\
		\hline
		V2\_02 & 0.057 &0.058 & -1.6\%\\
		\hline
		V2\_03 & x &x & x\\
		\hline
		\newline
	\end{tabular}
	
	\caption{Comparison between ORB-SLAM2 and DF-SLAM in EuRoC dataset with loop closure added.}
\end{table}

\begin{figure*}
	\centering
	\subfigure[DF-SLAM]{
		\fbox{
			\includegraphics[width=0.45\linewidth]{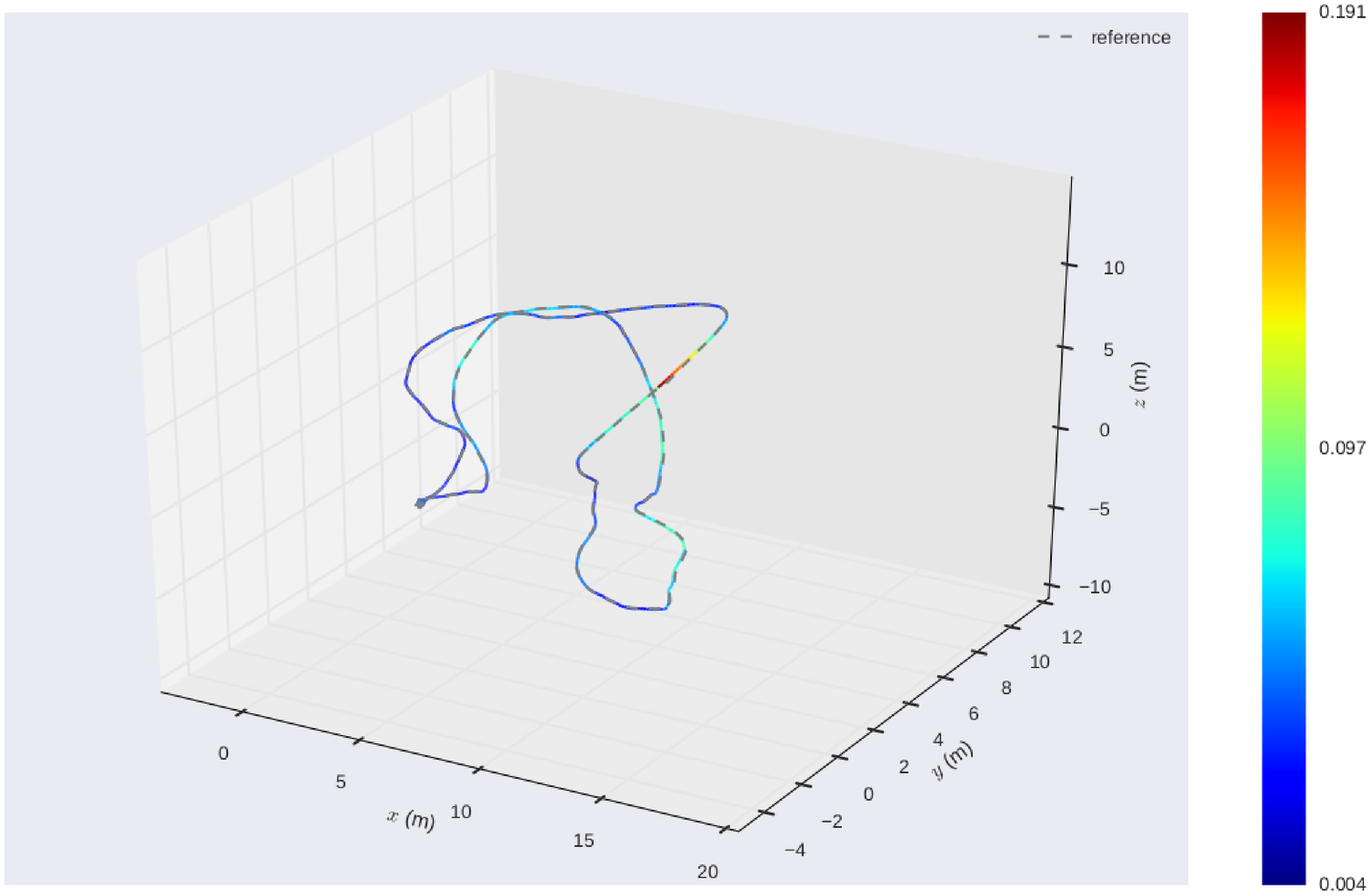}
		}
	}
	\subfigure[ORB-SLAM2]{
		\fbox{
			
			\includegraphics[width=0.45\linewidth]{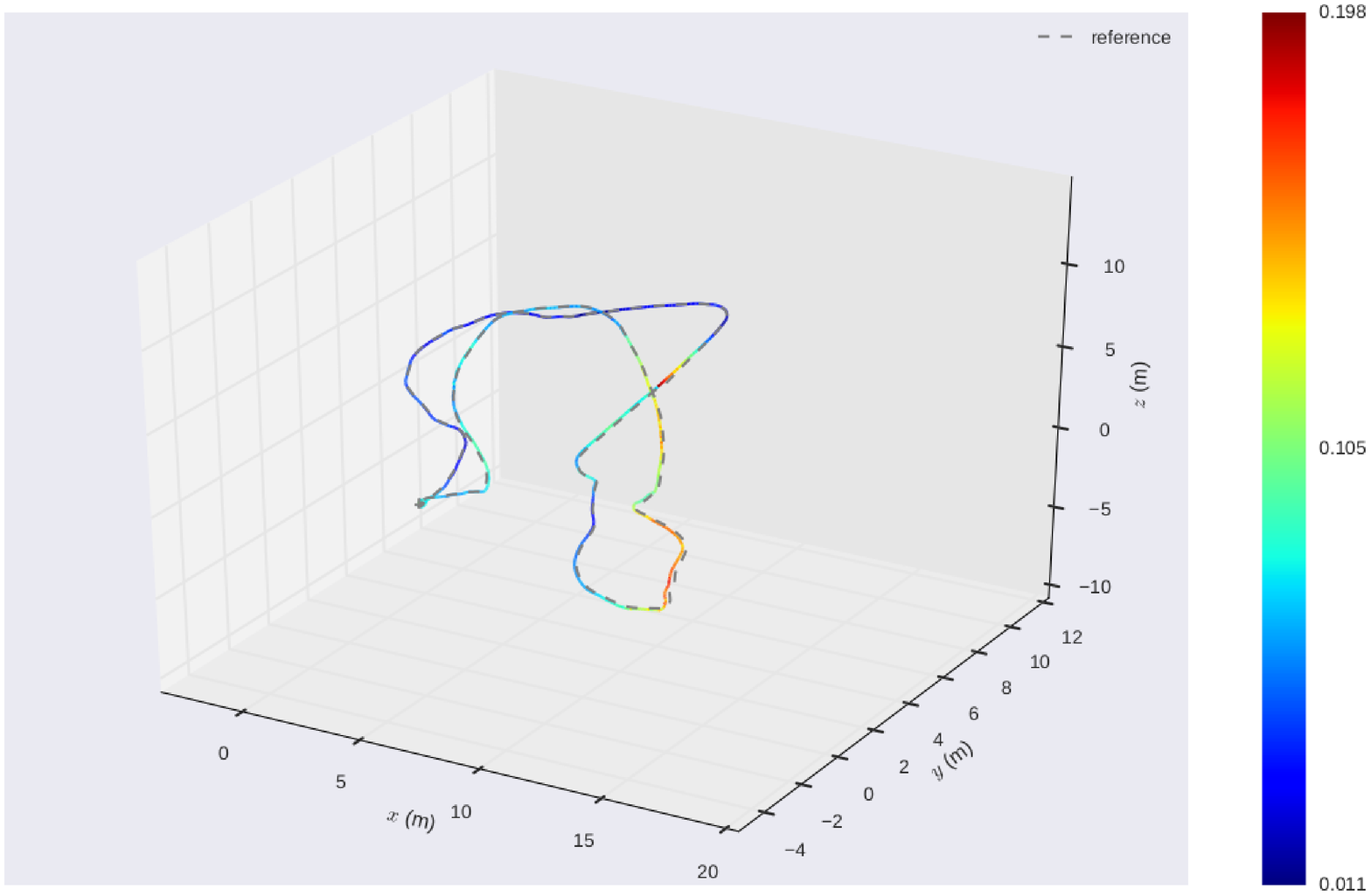}
		}
	}
	\caption{An example of MH 04 difficult sequence in EuRoC dataset. Above:DL-SLAM Below:ORB-SLAM2}
	\label{img}
\end{figure*}

\begin{table}[h]
	\centering
	\linespread{1}
	\begin{tabular}{cccc}
		\hline
		Dataset & ORB-SLAM2  & DF-SLAM & Improvement \\
		\hline
		MH\_01  & 0.038 & 0.036 & 4.3\%\\
		\hline
		MH\_02  & 0.047 & 0.050 & -8.3\%\\
		\hline
		MH\_03 & 0.039 & 0.043 & -10.6\%\\
		\hline
		MH\_04 & 0.147 & 0.060 & 58.9\%\\
		\hline
		MH\_05 & 0.059 &0.044 & 25.3\%\\
		\hline
		V1\_01 & 0.087 &0.086 & 1.5\%\\
		\hline
		V1\_02 & 0.097 &0.069 & 28.7\%\\
		\hline
		V1\_03 & 0.189 &0.114 & 39.8\%\\
		\hline
		V2\_01 & 0.071 &0.068 & 2.6\%\\
		\hline
		V2\_02 & 0.114 &0.102 & 10.1\%\\
		\hline
		V2\_03 & x &x & x\\
		\hline
		\newline
	\end{tabular}
	
	\caption{Comparison between ORB-SLAM2 and DF-SLAM in EuRoC dataset without loop closure.}
\end{table}
We evaluate the improved system in public EuRoC dataset, that consists of 11 sequences variant in scene complexity and sensor speed. The difficult sequences with intense lighting, motion blur, and low-texture areas are challenging for visual SLAM systems. As the ground truth of trajectory is provided in EuRoC, we use root-mean-square error(RMSE) for the representation of accuracy and stability. As we have mentioned above, we only change the threshold for feature matching and remain everything else the same as the original ORB-SLAM2 system, including the number of features we extract, time to insert a keyframe, ratio to do knn test during bow search period and so on. We also use the same pair of thresholds for each sequence.

We operate our system on each sequence for ten times and record both mean RMS errors for each data sequence and variance of these tests. As is illustrated in Figure 4, our method outperforms ORB-SLAM in MH sequences and perform no worse than ORB-SLAM in V sequences. Note that MH sequence is lack of loops and rely heavily on the performance of features while V sequence will always operate global pose optimization, we can easily find our method outstanding.

What is more, considering the variance of each test, we find that our system is quite stable no matter the situation. While the performance of ORB-SLAM2 may vary from time to time, we remain steady in each test we run.

We hold that the ability to walk a long way without much drift is a practical problem and matters a lot. We can never make sure that the environment we need to reconstruct is enough small and contains as many loops as we need to optimize our map. To further verify the performance of our system, we close the global bundle adjustment module(Loop Closing Thread) and repeat the test we run. We can easily find that our method outperforms ORB-SLAM2 at all V sequences, which has proved that DL-SLAM is actually more stable and accurate especially when the camera needs to go a long way without loops for global optimization.
\subsubsection*{B. TUM Dataset}
We further prove our robustness and accuracy on TUM Dataset, another famous dataset among SLAM researchers. The TUM dataset consists of several indoor object-reconstruction sequences. Since most of the sequences we used to make evaluation are captured by hand-holding cameras, these datasets contain terrible twitter from time to time. Such sequences are therefore excellent to test the robustness of our system.

We still use the same pair of features as in EuRoC datasets and other numerical features the same as ORB-SLAM2. We are happy to find that in TUM Datasets, where other SLAM systems lose their trajectory frequently, our system works well all the time. We take fr1/desk sequence as an example in Fig 7, where ORB-SLAM2 lost seven times at the same place in our entire ten tests and DF-SLAM covers the whole period easily. Similar to EuRoC, we find that DF-SLAM achieves much better results than ORB-SLAM2 among sequences that do not contain any apparent loops, and perform no worse that ORB-SLAM2 when there is no harsh noise or shake.
\begin{figure}
	\centering
	
	\subfigure[Track Lost]{
		
		\begin{minipage}[b]{0.5\textwidth}
			\fbox{
				\includegraphics[width=3.25in]{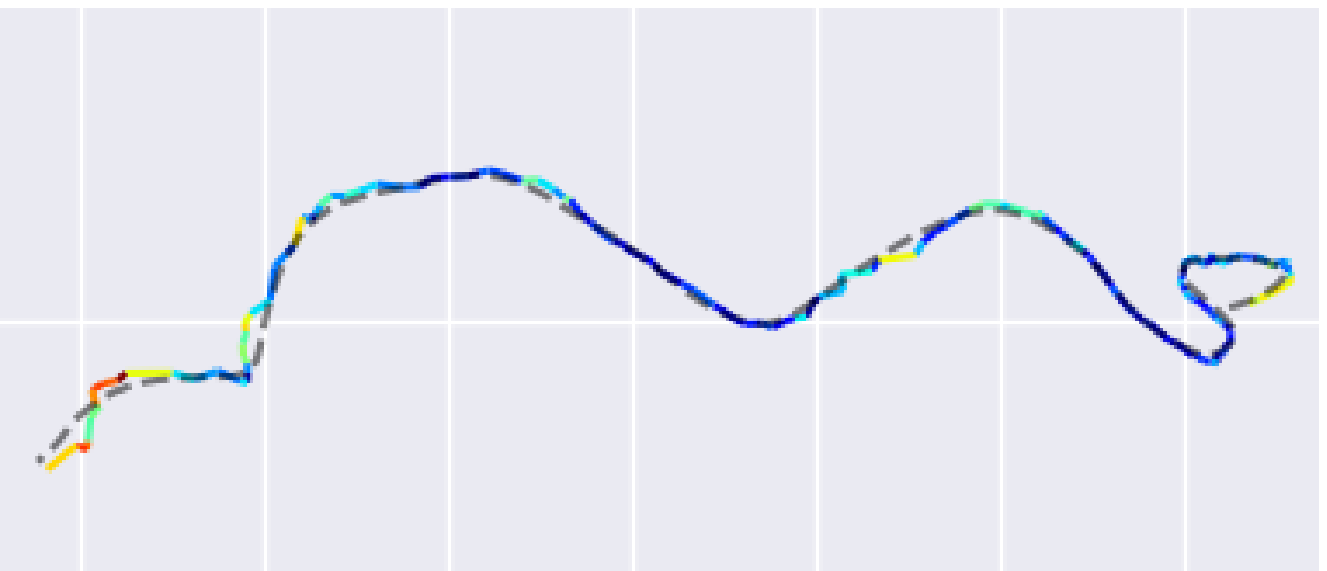}
			}
		\end{minipage}
	}
	\subfigure[Full Trajectory]{
		\begin{minipage}[b]{0.5\textwidth}
			\fbox{
				\includegraphics[width=3.25in]{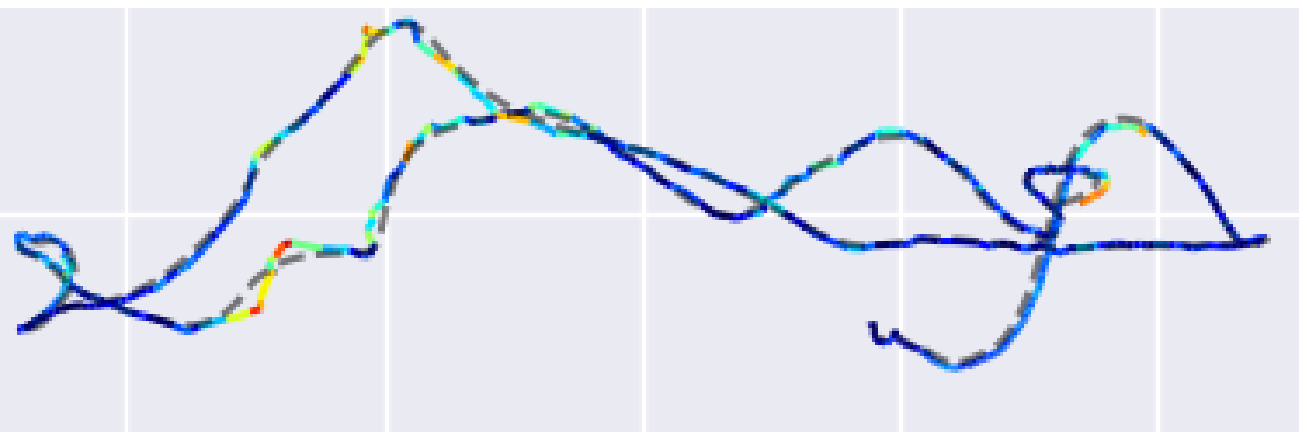}
			}
		\end{minipage}
	}
	
	\caption{Lost and Tracked sequence of fr1/desk. The camera starts from right to left and turn around to the right side.}
	\label{img}
\end{figure}

\begin{table}
	\centering
	\small
	\linespread{1}
	\begin{tabular}{ccccc}
		\hline
		Dataset & ORB-SLAM2  & DF-SLAM & Improvement & Tracked \\
		\hline
		fr1/desk  & 0.025 & 0.015 & 36.9\% & 3/10\\
		\hline
		fr1/desk2  & 0.028 & 0.021 & 24.5\% & 7/10\\
		\hline
		fr1/room & 0.058 & 0.041 & 28.9\% & 10/10\\
		\hline
		fr2/desk & 0.0089 & 0.0097 & -9.8\% & 10/10\\
		\hline
		fr2/xyz & 0.0038 &0.0030 & 19.5\% & 10/10\\
		\hline
		fr3/office & 0.011 &0.011 & -0.7\% & 10/10\\
		\hline
		fr3/nst & 0.022 &0.012 & 45.4\% & 10/10\\
		\hline
		
		\newline
	\end{tabular}
	
	\caption{Comparison between ORB-SLAM2 and DF-SLAM in TUM dataset. Tracked Numer is number of tests not lost in total 10 tests(ORB-SLAM2/DF-SLAM).}
\end{table}

\subsection{Runtime Evaluation}

We measure the run-time of the deep feature extraction using GeForce GTX TITAN X/PCIe/SSE2. A single forward pass of the model runs 7e-5 seconds for each patch based on pytorch c++ with CUDA support. The time spent on the feature extraction of one image is 0.09 seconds(1200 key points). Together with time to do tracking, mapping and loop closing in parallel, our system runs at a speed of 10 to 15fps. We find that since that our feature is much more robust and accurate, we can operate the whole system with a smaller number of features without losing our position. Therefore, there is still much space left for us to speed up the entire system and move forward to real-time.

\subsection{Local Feature Descriptor}
\begin{figure}[h]
	\centering
	\subfigure[The matching result on HPathes dataset.]{
		\begin{minipage}[b]{0.5\textwidth}
			\fbox{
				\includegraphics[width=3in]{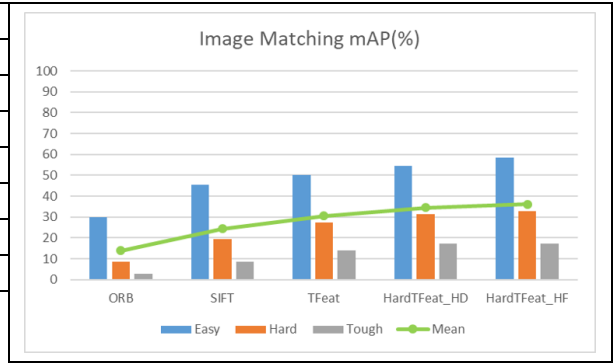}
			}
		\end{minipage}
	}
	\subfigure[The retrieval result on HPathes dataset.]{
		\begin{minipage}[b]{0.5\textwidth}
			\fbox{
				\includegraphics[width=3in]{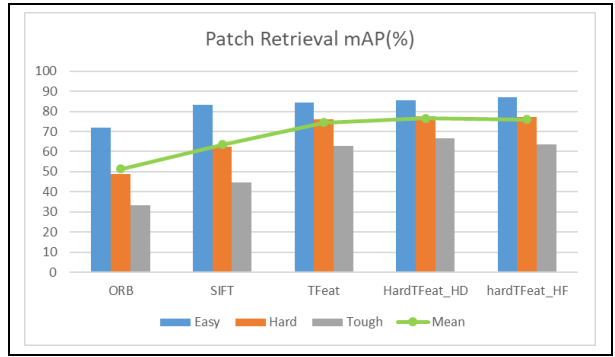}
			}
		\end{minipage}
	}
	\subfigure[The verification result on HPatches dataset.]{
		\begin{minipage}[b]{0.5\textwidth}
			\fbox{
				\includegraphics[width=3in]{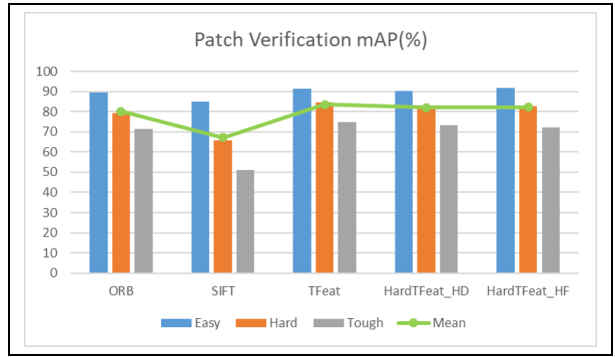}
			}
		\end{minipage}
	}
	\caption{The matching result on HPathes dataset. TFeat stands for the original TFeat network with simple training strategy. HardTFeat\_HD uses hard negative mining strategy, which is trained on original HPatches dataset. HardTFeat\_HF is the model we trained using FAST-based HPathces. }
	\label{img}
\end{figure}
Besides, we separately evaluate the performance of local feature descriptor that we used in DL-SLAM.

We use evenly distributed FAST detector to build the training dataset. All training is done using
pytorch and stochastic gradient descent solver with the learning
rate of 0.01, the momentum of 0.9 and weight decay of 0.0001.
We also use typical data augmentation techniques, such as
random rotation and crop, to improve the robustness of our
network.

We train our deep feature using different training strategies on HPatch training set and test them on testing set also provided by HPatch. We choose ORB and SIFT, two of the most popular descriptors as a comparison. Learned features outperform traditional ones in every task. Especially, HardTFeat\_HD shows a clear advantage over TFeat in matching function, which demonstrates the superiority of the strict hard negative mining strategy we use. HardTFeat\_HD and HardTFeat\_HF are trained on different datasets but show similar performance on both matching and retrieval tasks.


\section{Conclusion}
We propose DF-SLAM system that combines robust learned features with traditional SLAM techniques. DF-SLAM makes full use of the advantages of deep learning and geometric information and demonstrates outstanding improvements in efficiency and stability in numerous experiments. It can work stably and accurately even in challenging scenes. Our idea of making use of deep features provides better data associations and is an excellent aspect of doing further research on. The fantastic result proves the success of our novel idea that enhancing SLAM systems with small deep learning modules does lead to exciting results.

In future work, we will dedicate on the stability of DF-SLAM to handle difficult localization and mapping problems under extreme conditions. The speed of deep-learning-enhanced SLAM system is also within our consideration. What's more, we aim to design a robust local feature detector that matches the descriptors used in our system. Online learning is also an attractive choice to increase the modality of our system. We even decide to make use of global features to improve global bundle adjustment and establish a whole system for DL enhanced SLAM systems. We believe that such combination can figure out a great many non-geometric problems we are faced with and promote the development of SLAM techniques.

{\small
	\bibliographystyle{ieee}
	\bibliography{egbib}
}

\end{document}